\documentclass{article}

\usepackage{microtype}
\usepackage{graphicx}
\usepackage{subcaption}
\usepackage{booktabs} % for professional tables

\usepackage{hyperref}

\usepackage[preprint]{icml2026}

\usepackage{amsmath}
\usepackage{amssymb}
\usepackage{mathtools}
\usepackage{amsthm}
\usepackage[table]{xcolor}
\usepackage{multirow}
\usepackage{makecell}
\usepackage[capitalize,noabbrev]{cleveref}

\theoremstyle{plain}

\theoremstyle{definition}

\theoremstyle{remark}

\usepackage[textsize=tiny]{todonotes}

\icmltitlerunning{TEAM: Temporal–Spatial Consistency Guided Expert Activation for MoE Diffusion Language Model Acceleration}

\begin{document}

\twocolumn[
  % \icmltitle{Tailored Expert Activation: Exploiting Temporal–Spatial Consistency for Efficient MoE Diffusion Language Models}
  \icmltitle{TEAM: \underline{T}emporal–Spatial Consistency Guided \underline{E}xpert \underline{A}ctivation for \underline{M}oE Diffusion Language Model Acceleration}




  \begin{icmlauthorlist}
    \icmlauthor{Linye Wei}{AI,IC}
    \icmlauthor{Zixiang Luo}{YP}
    \icmlauthor{Pingzhi Tang}{AI,YP}
    \icmlauthor{Meng Li}{AI,IC,BAIC}
  \end{icmlauthorlist}

  \icmlaffiliation{AI}{Institute for Artificial Intelligence, Peking University, Beijing, China}
  \icmlaffiliation{IC}{School of Integrated Circuits, Peking University, Beijing, China}
  \icmlaffiliation{YP}{Yuanpei College, Peking University, Beijing, China}
  \icmlaffiliation{BAIC}{Beijing Advanced Innovation Center for Integrated Circuits, Beijing, China}

  \icmlcorrespondingauthor{Meng Li}{meng.li@pku.edu.cn}


  \vskip 0.3in
]



\printAffiliationsAndNotice{}  % no special notice (required even if empty)

\newcommand{\method}{TEAM}

\begin{abstract}
Diffusion large language models (dLLMs) have recently gained significant attention due to their inherent support for parallel decoding. Building on this paradigm, Mixture-of-Experts (MoE) dLLMs with autoregressive (AR) initialization have further demonstrated strong performance competitive with mainstream AR models. However, we identify a fundamental mismatch between MoE architectures and diffusion-based decoding. Specifically, a large number of experts are activated at each denoising step, while only a small subset of tokens is ultimately accepted, resulting in substantial inference overhead and limiting their deployment in latency-sensitive applications. In this work, we propose~\textbf{\method}, a plug-and-play framework that accelerates MoE dLLMs by enabling more accepted tokens with fewer activated experts. TEAM is motivated by the observation that expert routing decisions exhibit strong temporal consistency across denoising levels as well as spatial consistency across token positions. Leveraging these properties,~\method~employs three complementary expert activation and decoding strategies, conservatively selecting necessary experts for decoded and masked tokens and simultaneously performing aggressive speculative exploration across multiple candidates. Experimental results demonstrate that~\method~achieves up to 2.2× speedup over vanilla MoE dLLM, with negligible performance degradation. Code is released at \href{https://github.com/PKU-SEC-Lab/TEAM-MoE-dLLM}{https://github.com/PKU-SEC-Lab/TEAM-MoE-dLLM}.
\end{abstract}

\section{Introduction}
\label{sec:introduction}

Diffusion large language models (dLLMs) \cite{nie2025large,ye2025dream,khanna2025mercury} address limitations of autoregressive (AR) generation by adopting bidirectional attention, which enables parallelized decoding and positions dLLMs as a compelling alternative to conventional AR models. Recent advances \cite{wang2025diffusion,wu2025fast,fu2025efficient,liu2025wedlm} with AR initialization further strengthen this paradigm by incorporating strong autoregressive training priors while remaining compatible with KV cache, achieving both higher accuracy and better inference efficiency than AR models of comparable scale.

Thanks to the advantages in training and inference efficiency as well as their strong scalability, Mixture of Experts (MoE) \cite{shazeer2017outrageously,jiang2024mixtral} architectures have become dominant in state-of-the-art AR language models \cite{qwen3,liu2024deepseek,comanici2025gemini}. Initializing dLLMs from such models, like SDAR \cite{cheng2025sdar} and LLaDA 2.0 \cite{bie2025llada20scalingdiffusionlanguage}, further enhances the effectiveness of diffusion-based decoding, reinforcing the competitiveness of this paradigm compared to the latest AR LLMs \cite{dinfer,team2025every}.

Nevertheless, we observe that naively integrating MoE architectures into dLLMs can substantially degrade inference efficiency. During each diffusion iteration, all tokens within a decoding block are processed in parallel under bidirectional attention, with each token independently selecting its routed experts. However, only a small subset of tokens whose confidence exceeds a predefined threshold are ultimately unmasked. Due to the heterogeneity of expert routing decisions across tokens, a single forward pass typically activates a large fraction of the available experts, leading to significant memory access and communication overhead, while yielding only a limited number of accepted tokens.

\begin{figure}[ht]
  \vskip 0.1in
  \begin{center}
    \centerline{\includegraphics[width=\columnwidth]{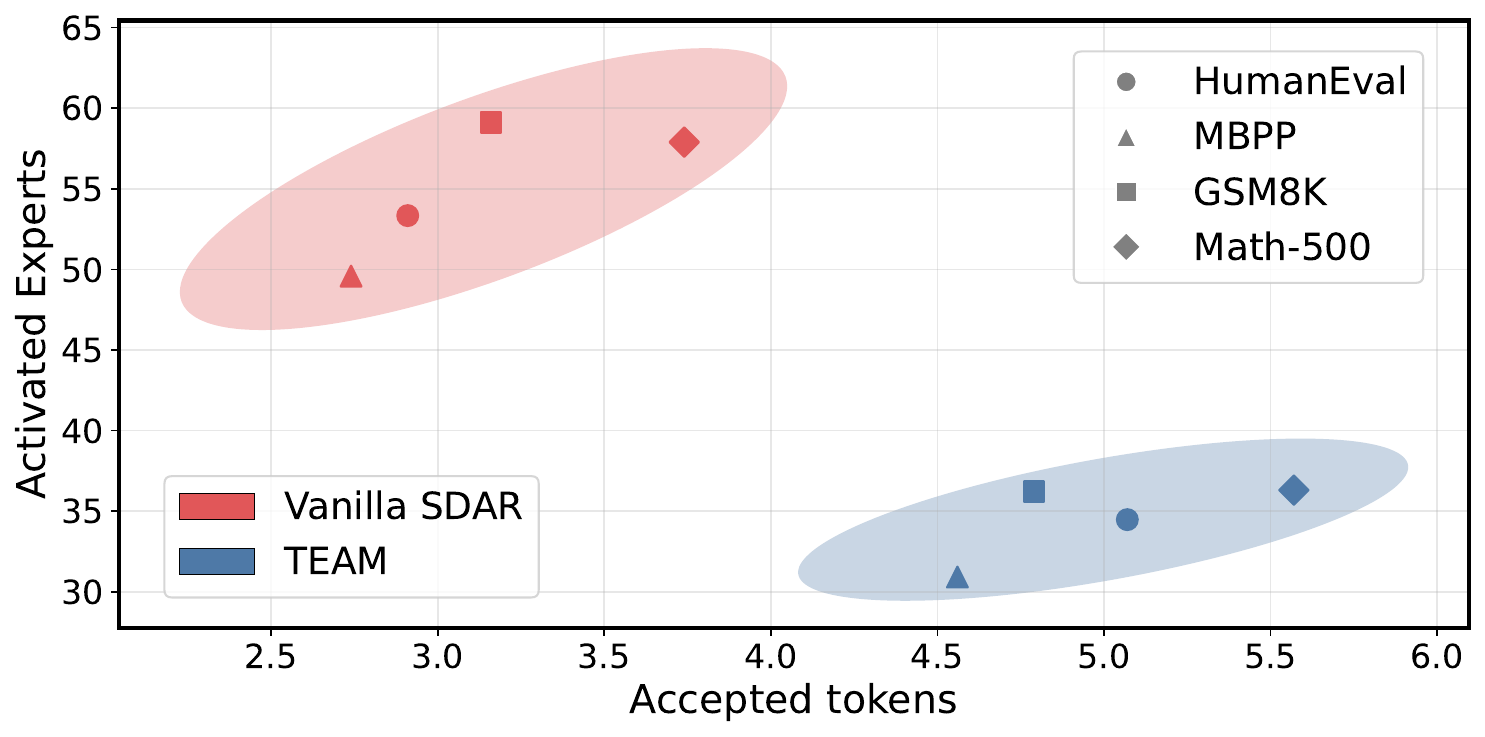}}
    \caption{
      Activated experts vs. accepted tokens per forward pass in SDAR 30B-A3B. TEAM decodes more tokens with fewer experts activated in an iteration.
    }
    \label{Fig1}
  \end{center}
\end{figure}

As illustrated in \cref{Fig1}, for SDAR \cite{cheng2025sdar} (8 experts per token), the observed ratio of activated distinct experts to ultimately accepted tokens is substantially higher than eight in practice. This phenomenon highlights the difficulty of applying MoE architectures to dLLMs: it effectively reverts to dense-model inefficiency. While this overhead can be amortized across concurrent requests in cloud deployments, it becomes a critical bottleneck in scenarios that are highly sensitive to decoding speed and tail latency, as well as on edge platforms with constrained hardware resources.

To address this challenge, we propose \textbf{\method}, which is developed based on our core observation that although both involve multi token decoding, block-wise inference in dLLMs is fundamentally different from multi batch inference in autoregressive models, exhibiting strong temporal and spatial consistency. \textbf{Along the temporal dimension}, dLLMs repeatedly process the same tokens within a block across multiple denoising levels, and for decoded tokens that have already been accepted, their values remain fixed in subsequent iterations within the block, yet they continue to participate in forward computation due to the bidirectional attention mechanism, leading to numerous unnecessary expert activations. \textbf{Along the spatial dimension}, the routing of unaccepted, masked tokens is highly concentrated, with relatively little variation in expert selection across tokens. Moreover, owing to the latent causal logic of the decoding process, the acceptance order exhibits spatial locality, suggesting that a substantial portion of masked tokens can be predicted to remain unaccepted in early iterations.

Building on these observations,~\method~implements three customized expert activation strategies for tokens within each block. For \textbf{decoded tokens}, we introduce a delayed caching mechanism, activating experts only for recently accepted tokens. For masked tokens, we further partition them into \textbf{hot tokens}, which are more likely to be accepted in the near future, and \textbf{cold tokens}, which are unlikely to be accepted. Exploiting the concentration of expert routing among masked tokens,~\method~performs speculative exploration on hot tokens to increase the acceptance rate, while rerouting cold tokens to experts that are already activated by decoded or hot tokens. Our main contributions can be summarized as follows:
\begin{itemize}
  \item We investigate the inefficiency of naively applying MoE architectures to dLLMs. To the best of our knowledge, this is the first study to specifically analyze expert activation characteristics in MoE dLLMs.
  \item Leveraging the temporal-spatial consistency of block-wise decoding in dLLMs, we propose~\method, which implements three complementary expert activation and decoding strategies, achieving more accepted tokens with fewer activated experts.
  \item Extensive experiments demonstrate that, with our tailored expert activation,~\method~achieves up to 2.2× speedup while preserving model performance.
\end{itemize}
\section{Related work}
\label{sec:related work}

\textbf{Diffusion Large Language Models (dLLMs).} 
Autoregressive large language models (LLMs) \cite{achiam2023gpt,grattafiori2024llama,qwen3} have demonstrated remarkable capabilities across a wide range of applications across a wide range of applications such as text generation, medical science \cite{zhu2025pathology}, and embodied AI \cite{yang2026efficientnav}, yet their inference efficiency is constrained by autoregressive decoding. Inspired by the success of diffusion in generative modeling \cite{ho2020denoising,yan2025entropy,yan2026less,zhu2025fmri2ges,yan2026pixel}, diffusion large language models (dLLMs) \cite{nie2025large,ye2025dream,li2025refusion,yang2026improving,feng2026dvoting} mitigate this limitation by enabling parallel decoding via bidirectional attention mechanism. In this paradigm, the entire response is represented as masked tokens, and all positions are decoded in each forward pass. Tokens whose confidence exceeds a predefined threshold are accepted, while the remaining tokens are re-masked and refined in subsequent iterations. However, the global bidirectional attention prevented reuse of KV cache, resulting in limited efficiency gains. More recent work \cite{wang2025diffusion,tian2025next,arriolablock,gong2025scaling} adopts a block diffusion strategy, partitioning the response into multiple blocks. By retraining from autoregressive initialization, these models learn to apply bidirectional attention within a block to enable parallel decoding, while using causal attention across blocks to support KV cache. In this way, dLLMs inherit strong autoregressive priors for accuracy while simultaneously improving inference efficiency through parallel decoding and cache reuse.

\textbf{Mixture-of-Experts (MoE).}
Unlike dense models which activate all parameters during each forward pass, Mixture-of-Experts (MoE) architectures \cite{shazeer2017outrageously,jiang2024mixtral,xu2026grouter} replace a single feedforward layer with a group of parallel expert networks and employ a gating mechanism to dynamically select a subset of experts for each token. By enabling expert specialization and sparse parameter activation, MoE architectures scale more effectively to achieve higher model capacity while improving both training and inference efficiency \cite{huang2025hd,zhong2025hybrimoe}, and have consequently become a dominant design in recent LLMs \cite{qwen3,liu2024deepseek,comanici2025gemini}. This paradigm has also been extended to dLLMs \cite{zhu2025llada,cheng2025sdar,bie2025llada20scalingdiffusionlanguage}. However, in dLLMs, all tokens within a block are processed at every diffusion iteration, and each token independently activates its routed experts. As a result, even when the batch size is one, a large fraction of the model’s parameters may be activated in a single forward pass. Such scenarios are often precisely those that are most sensitive to decoding latency, thereby limiting the practical deployment of MoE-based dLLMs.

\textbf{Acceleration of dLLMs.}
Several studies have explored techniques to accelerate dLLMs. Early approaches \cite{wu2025fast,liu2025dllm,ma2025dkv} reduce computation through approximate KV caching, but rely on fixed and coarse-grained intervals for cache refresh. Other methods \cite{chen2025dpad,song2025sparse,jiang2025d,qian2026d3llm} limit arithmetic intensity via sparsification, which becomes less critical for block diffusion. A separate line of work \cite{gao2025self,agrawal2025spiffy,wei2025orchestrating,wu2025free,zhu2025latent} integrates latent refinement or speculative decoding \cite{chen2023accelerating,leviathan2023fast} into dLLMs to further increase decoding parallelism. Beyond dense models, the strong empirical performance of MoE dLLMs has recently motivated efforts to accelerate this paradigm, such as dInfer \cite{dinfer}. However, dInfer primarily targets general dLLM acceleration and focuses only on expert-parallel execution for cloud-scale MoE deployment. In contrast, we present the first dedicated analysis of expert activation behavior in MoE dLLMs and propose TEAM, which exploits temporal–spatial consistency to tailor distinct expert activation strategies for different tokens, thereby improving decoding efficiency.
\section{Preliminary and Motivation}
\label{sec:Preliminary and motivation}

% \subsection{Preliminary and Motivation}
The inference process of a dLLM begins by initializing the response $Y$ with $N=B\times~L$ $\mathtt{[MASK]}$ tokens, where the response is partitioned into $B$ blocks, each containing $L$ tokens. The $i$-th block is denoted as $Y_i=\left[y_i^0, y_i^1, \cdots, y_i^{L-1}\right]$. Given a prompt $P$ , the model $p_{\vartheta}$ performs block-wise decoding, factorizing the response $\widehat{Y}$ as:

\begin{equation}
    \label{eq:dLLM_1}
    p_{\vartheta}(\widehat{Y} \mid P)=\prod_{i=1}^B p_{\vartheta}\left(\widehat{Y}_i \mid P, Y_{\leq i}\right)
\end{equation}

Concretely, within each block, the model iteratively samples from the $\mathtt{[MASK]}$ tokens. For the $i$-th block, a single forward pass produces token predictions and corresponding confidence scores, given by:

\begin{equation}
    \label{eq:dLLM_2}
    \begin{split}
    \widehat{y_i^k} &= \underset{v \in V}{\operatorname{argmax}} p_{\vartheta}\left(y_i^k=v \mid P, Y_{\leq i}\right) \text { and } \\
    c_k &= p_{\vartheta}\left(y_i^k=\widehat{y_i^k} \mid P, Y_{\leq i}\right), k \in[0,1, \cdots, \mathrm{~L}]
    \end{split}
\end{equation}

where $V$ denotes the model’s vocabulary. Only tokens whose confidence exceeds a predefined threshold $\tau$ are accepted, while the remaining tokens are re-masked for subsequent iterations:

\begin{equation}
    \label{eq:dLLM_3}
    y_i^k=\left\{
    \begin{array}{c@{\hspace{1em}}l}
    \widehat{y_i^k}, & \text{if } c_k>\tau \\
    {[\mathrm{MASK}]}, & \text{otherwise}
    \end{array}
    \right., k \in[0,1, \cdots, \mathrm{L}]
\end{equation}

Once all positions within a block are unmasked, the block is cached and decoding proceeds to the next, repeating this process until the end-of-sequence $\mathtt{[EOS]}$ token is generated.

\begin{figure*}[ht]
  \vskip 0.1in
  \begin{center}
    \centerline{\includegraphics[width=0.98\linewidth]{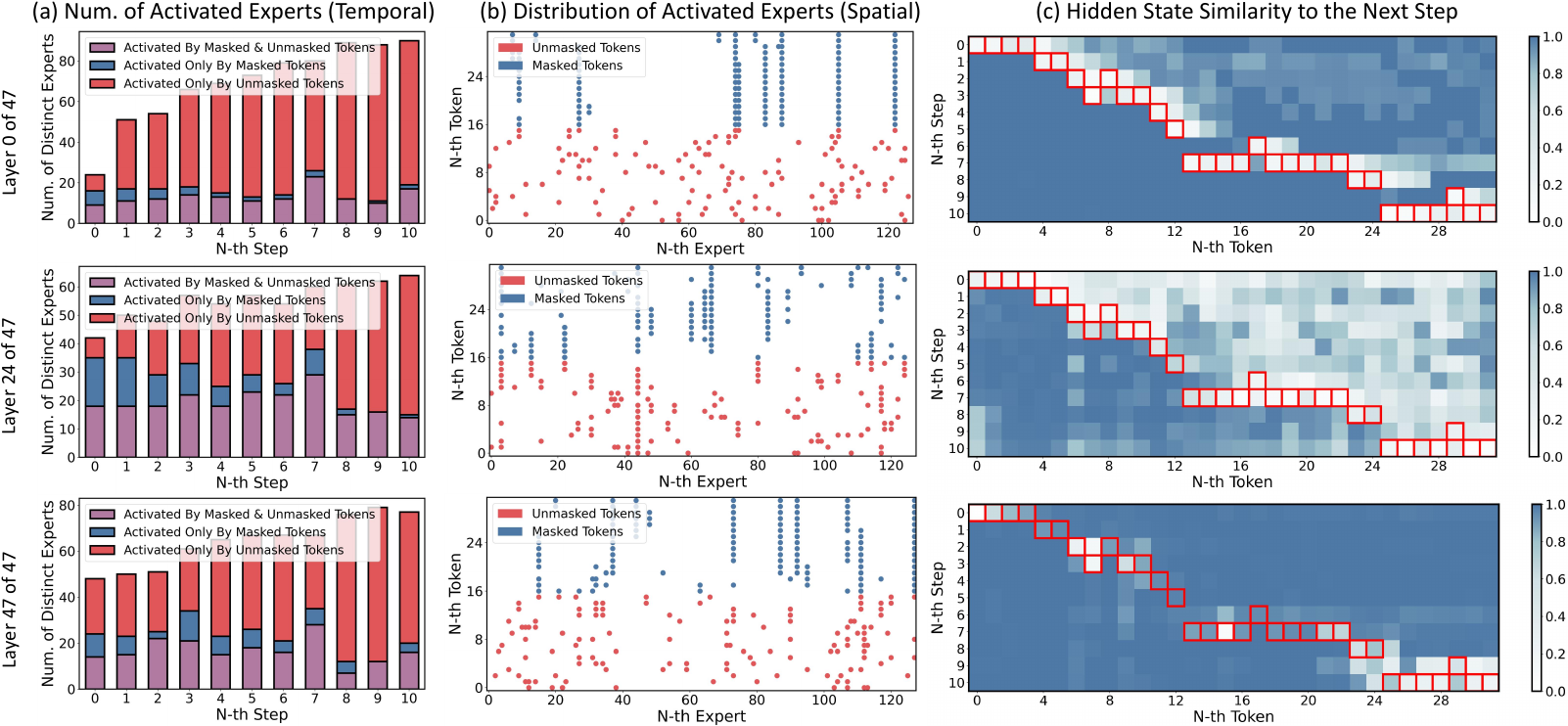}}
    \caption{
      Temporal-spatial characteristics of expert activation and decoding with the SDAR 30B-A3B model on a prompt from the GSM8K dataset. Results are shown for layers 0, 24, and 47 (of 47). (a) Number of activated experts across decoding iterations. (b) Distribution of experts activated by decoded and masked tokens at step 6 (of 11). (c) Token acceptance positions at each iteration, together with hidden state similarity relative to the subsequent iteration. 
      % The generality across more datasets is shown in Appendix~\ref{sec:Motivation}.
    }
    \label{Fig2}
  \end{center}
\end{figure*}

The decoding efficiency of dLLMs arises from the parallel processing of all tokens within a block, which allows multiple tokens to be accepted in a single iteration. However, when this block-wise decoding paradigm is combined with MoE architectures, the parallel tokens collectively activate a large fraction of the experts, negating the benefits of sparse parameter activation. As a result, the combined system can underperform relative to the theoretical gains of its constituent components. To better understand this inefficiency, we analyze the decoding trajectory of a single block along with its associated expert activation patterns, as illustrated in \cref{Fig2}, from which we derive several key observations.

\textbf{Temporal Consistency.}
Block-wise decoding in diffusion large language models (dLLMs) requires repeatedly processing the same block across successive denoising iterations, during which tokens are gradually accepted and propagated to subsequent steps. Although these accepted tokens no longer change and merely provide context for decoding the remaining masked tokens, they still incur full computation at every iteration and independently trigger expert activations in MoE layers.~\cref{Fig2}(a) reports the number of experts activated across iterations at three representative layers (first, middle, and last). This repeated computation on already decoded tokens leads to substantial additional expert activations at each step. As decoding progresses and more tokens become fixed, the fraction of total activated experts attributable to these tokens increases, ultimately dominating expert activation in later iterations.

\textbf{Spatial Consistency.}
In contrast to the token-specific activations of decoded tokens, variability among masked tokens primarily stems from positional encodings in their input embeddings. Consequently, spatially adjacent masked tokens exhibit highly consistent expert routing patterns across layers. As shown in~\cref{Fig2}(b), while decoded tokens activate a diverse set of experts with an approximately uniform distribution across candidates, masked tokens tend to concentrate their routing decisions on a small subset of experts. This observation suggests that a specific group of experts dominates the decoding of nearly all masked tokens, whereas experts outside this subset contribute only marginally or are invoked by very few tokens. The spatial concentration of expert activation also explains why, in~\cref{Fig2}(a), the experts activated by masked tokens constitute only a limited fraction of the total activated experts.

\textbf{Temporal-Spatial Locality.}
We further analyze the step-wise similarity of hidden states produced by each layer and mark the token positions that are unmasked at each iteration (highlighted by red boxes), as illustrated in~\cref{Fig2}(c). We find that the most significant change in the hidden state of a token occurs precisely between the iteration in which it is accepted and the subsequent iteration, consistent with observations in prior work \cite{ma2025dkv,song2025sparse,li2025sparse}. Once a token has been accepted and processed through one additional forward pass, its representation can be regarded as approximately stable. Moreover, the acceptance order within a block largely follows an autoregressive trend over time, and the tokens accepted at each iteration exhibit spatial clustering. This behavior is expected, given that dLLMs are initialized from autoregressive models and that natural language generation inherently follows a causal structure. As a result, tokens newly accepted at a given step tend to lie close to previously decoded tokens, while some tokens are unlikely to be accepted during early iterations.

\begin{figure*}[ht]
  \vskip 0.15in
  \begin{center}
    \centerline{\includegraphics[width=0.92\linewidth]{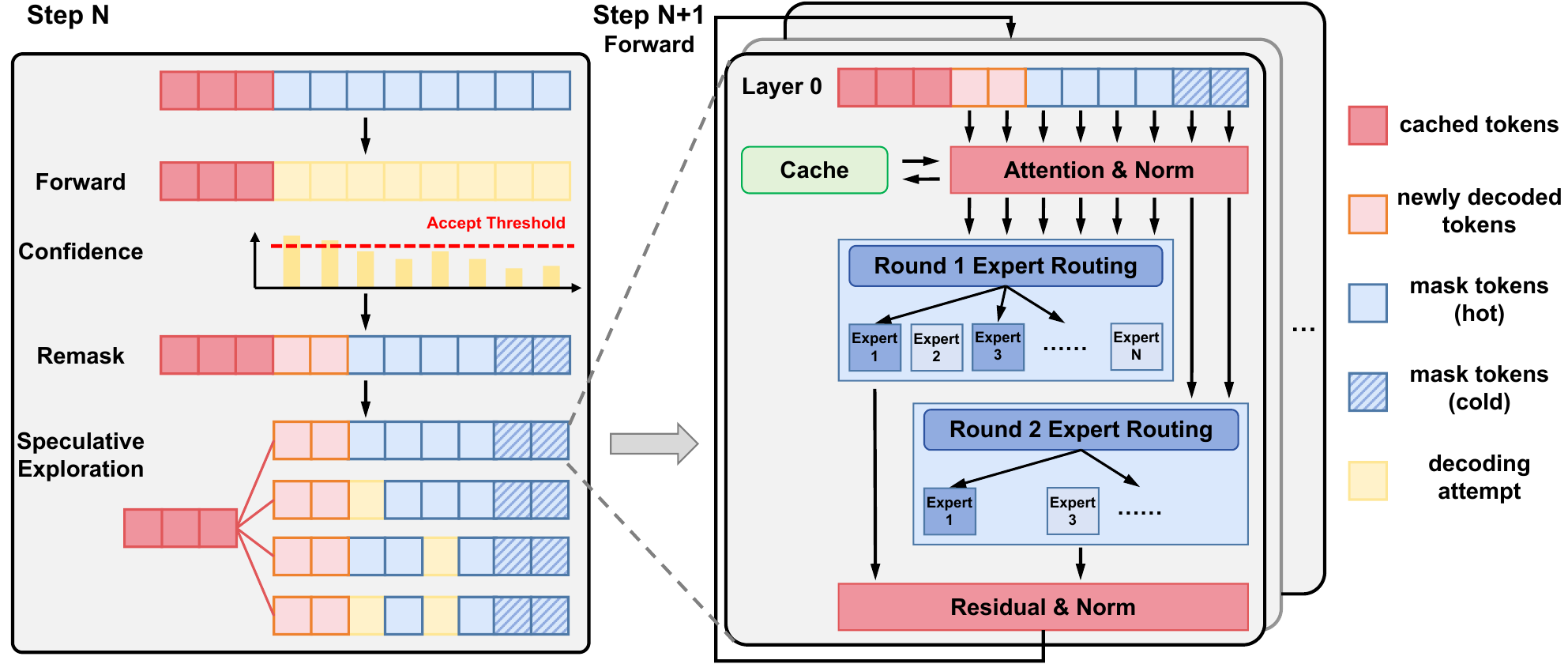}}
    \caption{
      Overview of our proposed \textbf{TEAM}. We apply differentiated expert activation and decoding strategies to tokens within each block. For \textbf{decoded tokens}, redundant computation is reduced through one-step delayed caching. For \textbf{mask tokens (hot)}, we adopt aggressive multi-branch speculative exploration to exploit idle compute resources and increase the token acceptance rate. For \textbf{mask tokens (cold)}, a double-round routing mechanism is introduced to constrain unnecessary expert activations.
    }
    \label{Fig3}
  \end{center}
  \vskip 0.14in
\end{figure*}

Building on these insights, we propose TEAM, which implements three complementary expert activation and decoding strategies tailored to the tokens within each block, as shown in~\cref{Fig3}.

\section{TEAM Methodology}
\label{sec:TEAM methodology}

\subsection{Delayed Caching for Decoded Tokens (DCD)}
As discussed before, once decoded tokens are incorporated as input and processed by an additional forward pass after being accepted, their hidden representations become approximately stable. This observation naturally motivates caching these tokens to avoid redundant computation across iterations. Specifically, at each iteration, computation is performed only for the masked tokens and the tokens newly accepted in the previous step, while the key–value pairs of tokens decoded in earlier iterations are reused from cache. After each forward pass, the KV pairs corresponding to newly accepted tokens are inserted into the cache and reused in subsequent iterations. 

A related strategy has been explored in prior work dKV-Cache \cite{ma2025dkv} targeting global bidirectional attention. To mitigate KV drift of decoded tokens under bidirectional attention, dKV-Cache additionally performs periodic global cache refresh by recomputing all tokens every $N$ iterations. This mechanism becomes less effective in dLLMs that adopt the block diffusion paradigm with native support for KV cache. In this setting, decoding no longer requires global token processing, and parallel computation is confined to tokens within a single block. Such block-level parallelism typically does not reach the compute–bandwidth balance point of modern hardware platforms such as GPUs, leaving decoding largely memory-bound and rendering fine-grained KV caching within a block unnecessary. 

However, as observed earlier, under MoE architectures, decoded tokens activate a large set of experts that are largely distinct from those activated by masked tokens, substantially increasing parameter activation density and memory access. This property makes caching decoded tokens particularly beneficial in MoE-based dLLMs. Moreover, by leveraging autoregressive priors and the near-autoregressive acceptance order during decoding, our delayed caching mechanism eliminates the need for periodic global cache refresh. As demonstrated in Section~\ref{sec:experiments}, this design improves decoding efficiency without sacrificing model quality.

\subsection{Speculative Exploration for Hot Tokens (SEH)}
Beyond eliminating redundant expert activations arising from repeatedly processing decoded tokens, we observe that expert activation and decoding for masked tokens are also inefficient. First, each activated expert is typically responsible for only a small number of tokens, resulting in frequent memory accesses with low computational intensity. Consequently, available compute resources are underutilized, increasing overall decoding latency. Second, the near-autoregressive decoding order and the spatial clustering of accepted tokens indicate that a subset of tokens has a low probability of being accepted in early iterations. Activating experts for such tokens only to remask them afterward leads to wasted computation. 

These observations motivate a differentiated computation strategy for masked tokens: we aim to improve decoding efficiency for tokens that are likely to be accepted in the near future, which we refer to as \textbf{hot tokens}, while reducing the computational overhead for tokens that are unlikely to be accepted, referred to as \textbf{cold tokens}. We identify two characteristics that make masked tokens more likely to be accepted in the next iteration: (1) their decoding attempt $y_i^k$ at the current iteration yields a relatively high confidence score $c_k$, even if it does not yet exceed the acceptance threshold; and (2) they are spatially closer to decoded tokens, and thus benefit from more informative contextual guidance. Formally, the hot tokens are defined as:

\begin{equation}
    \label{eq:dLLM_4}
    y_i^{k-h o t}=\left\{y_i^k \mid\left(c_k>\tau_h\right) \text { or }\left(\forall j,|k-j|<L_h\right)\right\}
\end{equation}

where $\tau_h$ denotes the confidence threshold to identify hot tokens, $j$ indexes the positions of currently decoded tokens, and $L_h$ specifies the maximum allowable distance from decoded tokens for a masked token to be classified as hot.

\begin{figure}[t]
  \vskip 0.1in
  \begin{center}
    \centerline{\includegraphics[width=\columnwidth]{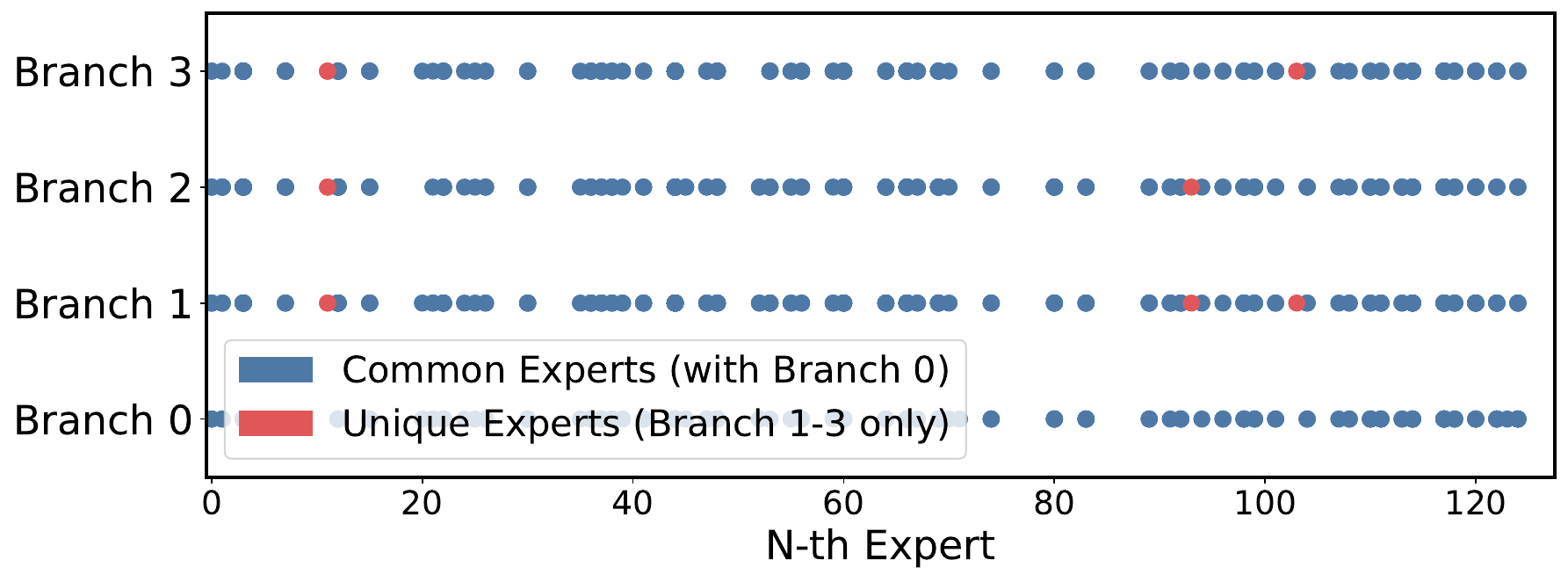}}
    \caption{
      Expert activation with speculative exploration in SDAR for a response from the GSM8K dataset, measured at layer 24 (of 47). 
      % More cases are provided in Appendix~\ref{sec:SEH}.
    }
    \label{Fig4}
  \end{center}
\end{figure}

For these hot tokens, we employ speculative exploration by additionally accepting the tokens with the top-$k$ confidence scores to construct multiple branches. By decoding and verifying these branches in parallel, we increase the acceptance rate at each iteration, thereby reducing the total number of decoding steps and overall latency. Under bidirectional attention, modifying any single token affects all tokens within the block, implying that each additional candidate effectively incurs the full computational cost of an entire block. For dense models, this behavior is prohibitive, as it rapidly increases computational intensity and shifts the decoding bottleneck from memory-bound to compute-bound, often requiring multi-GPU parallelism to achieve meaningful speedups \cite{xu2025lopa}. In contrast, for MoE architectures, inference latency is dominated by feedforward layers, and computation is naturally distributed across experts. As illustrated in~\cref{Fig4}, introducing additional decoding branches increases the originally low arithmetic intensity of each expert, while the high similarity across branches avoids activating a large number of new experts. Consequently, our exploration remains effective for MoE architectures even on a single GPU.

\subsection{Limited Activation for Cold Tokens (LAC)}
Cold tokens are defined as masked tokens that are both distant from decoded token positions and associated with low confidence scores in previous decoding attempts. Such tokens are unlikely to be accepted in subsequent iterations, implying that activating experts uniquely routed to them is often unnecessary. In practice, decoding attempts for cold tokens have a high probability of being re-masked after each forward pass, rendering their dedicated expert activations largely wasteful.

Leveraging the spatial consistency, masked tokens tend to route to a largely shared subset of experts, with only minor variation across tokens. Based on this property, we apply a limited activation strategy for cold tokens, as summarized in~\cref{alg:Limited Activation}. Specifically, we first perform expert routing for newly accepted tokens that have not yet been cached, as well as for hot tokens, since these tokens require accurate expert activation to ensure decoding correctness. The union of experts selected in this stage defines a necessary expert set. We then conduct a second round of routing for cold tokens, restricting their activation to this expert set. Through this double-round routing mechanism, expert activation for cold tokens is strictly confined to necessary experts, avoiding the introduction of token-specific expert activations while preserving model quality and retaining the possibility that cold tokens may still be unexpectedly accepted in later iterations.

% \begin{algorithm}[tb]
%   \caption{Limited Activation for Cold Tokens}
%   \label{alg:Limited Activation}
%   \begin{algorithmic}
%     \STATE {\bfseries Input:} data $x_i$, size $m$
%     \REPEAT
%     \STATE Initialize $noChange = true$.
%     \FOR{$i=1$ {\bfseries to} $m-1$}
%     \IF{$x_i > x_{i+1}$}
%     \STATE Swap $x_i$ and $x_{i+1}$
%     \STATE $noChange = false$
%     \ENDIF
%     \ENDFOR
%     \UNTIL{$noChange$ is $true$}
%   \end{algorithmic}
% \end{algorithm}

\begin{algorithm}[tb]
   % \caption{Limited Activation (Appendix~\ref{sec:LAC})}
   \caption{Limited Activation for cold tokens}
   \label{alg:Limited Activation}
\begin{algorithmic}
   \STATE {\bfseries Input:} Decoded tokens $D$, Mask tokens $M$, Experts $E_0$
   \STATE {\bfseries Output:} Activated Experts $E_A$, Routing Weights $W$
   
   \STATE \textbf{// 1. Classfication of tokens}
   \STATE Find newly accepted tokens $D_a \subseteq D$
   \STATE Find hot tokens $H \subseteq M$ via \textbf{Eq.~\ref{eq:dLLM_4}}
   \STATE Define cold tokens $C \leftarrow M \setminus H$
   
   \STATE \textbf{// 2. First-round Routing}
   \STATE Necessary activation $W_1 \leftarrow Router(D_a, H, E_0)$
   \STATE Necessary experts $E_A \leftarrow \text{top-}k\;(W_1)$
   
   \STATE \textbf{// 3. Second-round Routing}
   \STATE Activation for cold tokens $W_2 \leftarrow Router(C, E_A)$
   \STATE Routing Weights $W \leftarrow Concat(W_1, W_2)$
   
   \STATE {\bfseries Return} $E_A, W$
\end{algorithmic}
\end{algorithm}
\section{Experiments}
\label{sec:experiments}

\subsection{Experimental Setup}
\begin{table*}[t]
    \centering
    % \vspace{0.1in}
    \caption{\textbf{Performance of TEAM on SDAR.} APF denotes the number of Activated experts Per Forward pass, TPF denotes accepted Tokens Per Forward pass, and APT denotes the equivalent number of Activated experts Per decoded Token.}
    % \vspace{0.1in}
    \label{SDAR Performance}
    \begin{tabular}{ccccccc}
        \toprule
        \textbf{Benchmark} & \textbf{Method} & \textbf{Score$\uparrow$} & \textbf{APF$\downarrow$} & \textbf{TPF$\uparrow$} & \textbf{APT$\downarrow$} & \textbf{Speedup} \\
        \midrule
        
        % Group 1: HumanEval
        \multirow{2}{*}{\makecell[c]{HumanEval\\\textit{\small(0-shot)}}} 
        & Vanilla & 79.27 & 53.34 & 2.91 & 18.33 & 1$\times$ \\
        & \cellcolor{gray!15}TEAM & \cellcolor{gray!15}79.88 (\textcolor{green!50!black}{+0.61}) & \cellcolor{gray!15}34.48 (\textcolor{green!50!black}{35\%$\downarrow$}) & \cellcolor{gray!15}5.07 (\textcolor{green!50!black}{1.74$\times$}) & \cellcolor{gray!15}6.80 (\textcolor{green!50!black}{63\%$\downarrow$}) & \cellcolor{gray!15}2.20$\times$ \\
        
        % Group 2: MBPP
        \multirow{2}{*}{\makecell[c]{MBPP\\\textit{\small(0-shot)}}} 
        & Vanilla & 65.76 & 49.59 & 2.74 & 18.10 & 1$\times$ \\
        & \cellcolor{gray!15}TEAM & \cellcolor{gray!15}65.76 (\textcolor{green!50!black}{+0.00}) & \cellcolor{gray!15}30.92 (\textcolor{green!50!black}{38\%$\downarrow$}) & \cellcolor{gray!15}4.56 (\textcolor{green!50!black}{1.66$\times$}) & \cellcolor{gray!15}6.78 (\textcolor{green!50!black}{63\%$\downarrow$}) & \cellcolor{gray!15}2.08$\times$ \\
        
        % Group 3: GSM8K
        \multirow{2}{*}{\makecell[c]{GSM8K\\\textit{\small(0-shot)}}} 
        & Vanilla & 90.60 & 59.11 & 3.16 & 18.71 & 1$\times$ \\
        & \cellcolor{gray!15}TEAM & \cellcolor{gray!15}90.30 
        (\textcolor{red!60!black}{--0.30}) & \cellcolor{gray!15}36.20 (\textcolor{green!50!black}{39\%$\downarrow$}) & \cellcolor{gray!15}4.79 (\textcolor{green!50!black}{1.52$\times$}) & \cellcolor{gray!15}7.56 (\textcolor{green!50!black}{60\%$\downarrow$}) & \cellcolor{gray!15}1.83$\times$ \\
        
        % Group 4: Math-500
        \multirow{2}{*}{\makecell[c]{Math-500\\\textit{\small(0-shot)}}} 
        & Vanilla & 76.00 & 57.90 & 3.74 & 15.48 & 1$\times$ \\
        & \cellcolor{gray!15}TEAM & \cellcolor{gray!15}75.40 
        (\textcolor{red!60!black}{--0.60}) & \cellcolor{gray!15}36.31 (\textcolor{green!50!black}{37\%$\downarrow$}) & \cellcolor{gray!15}5.57 (\textcolor{green!50!black}{1.49$\times$}) & \cellcolor{gray!15}6.52 (\textcolor{green!50!black}{58\%$\downarrow$}) & \cellcolor{gray!15}1.64$\times$ \\
        \midrule
        
        % Group 5: Average
        \multirow{2}{*}{Average} 
        & Vanilla & 77.91 & 54.99 & 3.14 & 17.66 & 1$\times$ \\
        & \cellcolor{gray!15}TEAM & \cellcolor{gray!15}77.84
        (\textcolor{red!60!black}{--0.07}) & \cellcolor{gray!15}34.48 (\textcolor{green!50!black}{37\%$\downarrow$}) & \cellcolor{gray!15}5.00 (\textcolor{green!50!black}{1.59$\times$}) & \cellcolor{gray!15}6.92 (\textcolor{green!50!black}{61\%$\downarrow$}) & \cellcolor{gray!15}1.94$\times$ \\
        
        \bottomrule
    \end{tabular} 
    \vspace{0.1in}
\end{table*}

\begin{table*}[t]
    \centering
    % \vspace{0.1in}
    \caption{\textbf{Sensitivity Analysis of Hot-Token Hyperparameters}. Model accuracy and the number of Activated experts Per Forward pass (APF) with DCD and LAC strategy on code generation benchmarks.}
    % \vspace{0.1in}
    \label{hyperparam_sensitivity}
    \setlength{\tabcolsep}{3.5pt}
    \begin{tabular}{ccccccccccc}
        \toprule
        \multirow{2}{*}{\textbf{Benchmark}} & 
        \multicolumn{2}{c}{\textbf{$\tau_h=0.4, L_h=6$}} & 
        \multicolumn{2}{c}{\textbf{$\tau_h=0.5, L_h=5$}} & 
        \multicolumn{2}{c}{\textbf{$\tau_h=0.6, L_h=4$}} & 
        \multicolumn{2}{c}{\textbf{$\tau_h=0.7, L_h=3$}} & 
        \multicolumn{2}{c}{\textbf{$\tau_h=0.8, L_h=2$}} \\
        \cmidrule(lr){2-3} \cmidrule(lr){4-5} \cmidrule(lr){6-7} \cmidrule(lr){8-9} \cmidrule(lr){10-11}
         & \textbf{Score} & \textbf{APF} & \textbf{Score} & \textbf{APF} & \textbf{Score} & \textbf{APF} & \textbf{Score} & \textbf{APF} & \textbf{Score} & \textbf{APF} \\
        \midrule
        HumanEval & 78.05 & 24.49 & 81.09 & 24.38 & 79.27 & 24.01 & 79.27 & 23.08 & 77.44 & 22.26 \\
        MBPP & 66.93 & 22.21 & 65.76 & 22.15 & 68.09 & 22.12 & 66.93 & 21.66 & 62.65 & 20.40 \\
        \midrule
        Average & 72.49 & 23.35 & 73.43 & 23.27 & 73.68 & 23.07 & 73.10 & 22.37 & 70.05 & 21.33 \\
        \bottomrule
    \end{tabular}
    \vspace{0.1in}
\end{table*}

\begin{table*}[t]
    \centering
    % \vspace{0.1in}
    \caption{\textbf{Performance with Delayed Caching for Decoded Tokens (DCD).} Refresh-4 denots block cache refresh every 4 steps, Refresh-8 denots block cache refresh every 8 steps, and Refresh-free denotes never refresh (Ours).}
    % \vspace{0.1in}
    \label{refresh_comparison}
    \setlength{\tabcolsep}{5pt}
    \begin{tabular}{cccccccccc}
        \toprule
        \multirow{2}{*}{\textbf{Benchmark}} & \multicolumn{3}{c}{\textbf{Refresh-4}} & \multicolumn{3}{c}{\textbf{Refresh-8}} & \multicolumn{3}{c}{\textbf{Refresh-free}} \\
        \cmidrule(lr){2-4} \cmidrule(lr){5-7} \cmidrule(lr){8-10}
         & \textbf{Score} & \textbf{APF} & \textbf{Speedup} & \textbf{Score} & \textbf{APF} & \textbf{Speedup} & \textbf{Score} & \textbf{APF} & \textbf{Speedup} \\
        \midrule
        HumanEval & \textbf{79.88} & 32.67 & 1.38$\times$ & 78.66 & 29.61 & 1.47$\times$ & \textbf{79.88} & 26.72 & 1.58$\times$ \\
        MBPP & \textbf{66.15} & 29.89 & 1.32$\times$ & \textbf{66.15} & 26.91 & 1.44$\times$ & 65.76 & 23.67 & 1.55$\times$ \\
        GSM8K & 90.52 & 35.45 & 1.27$\times$ & \textbf{90.60} & 31.23 & 1.35$\times$ & 90.45 & 27.99 & 1.44$\times$ \\
        Math-500 & \textbf{74.80} & 35.25 & 1.17$\times$ & 73.00 & 31.18 & 1.25$\times$ & 74.20 & 27.52 & 1.32$\times$ \\
        \midrule
        Average & 77.84 & 33.32 & 1.29$\times$ & 77.10 & 29.73 & 1.38$\times$ & 77.57 & 26.48 & 1.47$\times$ \\
        \bottomrule
    \end{tabular}
    \vspace{0.1in}
\end{table*}
  
Our experiments are primarily conducted on SDAR 30B-A3B \cite{cheng2025sdar}, a representative MoE-based diffusion language model that follows the block diffusion paradigm. We adopt the official open-sourced evaluation protocol provided by SDAR as our baseline. We note that for another MoE-based dLLM, LLaDA 2.0 \cite{bie2025llada20scalingdiffusionlanguage}, an official Hugging Face–format evaluation pipeline is not available, and therefore it is not used as the primary experimental platform in this study. To demonstrate the generality of our proposed TEAM across diverse tasks, we evaluate performance on two code generation benchmarks, HumanEval \cite{humaneval} and MBPP \cite{austin2021program}, as well as two mathematical reasoning benchmarks, GSM8K \cite{cobbe2021training} and Math-500 \cite{lightman2023let}. All experiments are conducted on an NVIDIA A100 80GB GPU.

Unless otherwise specified, we follow the official implementations of SDAR and LLaDA 2.0. The acceptance threshold for unmasking tokens is set to $\tau=0.95$, and the block size, which is the number of tokens decoded in parallel at each iteration, is fixed to 32. For masked token classification in TEAM, we identify hot tokens either by a confidence threshold $\tau_h=0.7$ or by requiring them to be no more than $L_h=3$ positions away from decoded tokens. In speculative exploration for hot tokens, the number of parallel branches is set to 4.

\subsection{Main Results}
We report the improvements of TEAM over the vanilla model in terms of expert activation and decoding efficiency, as summarized in~\cref{SDAR Performance}. Without TEAM, each layer of the vanilla model activates more than 50 experts on average per forward pass, approaching half of the total 128 experts. However, each forward pass accepts and unmasks only approximately three tokens, leading to more than twice the nominal routing cost (8 experts per token) for decoding a single token, and in the worst case up to 18 activated experts per decoded token. This reveals a fundamental inefficiency: although parallel decoding in dLLMs and sparse parameter activation in MoE architectures are both individually designed to be inference efficient, their naive combination becomes counterproductive. This inherent incompatibility significantly degrades the overall decoding speed.

In contrast, TEAM achieves decoding of more tokens with substantially fewer expert activations through its carefully designed expert activation and decoding strategies, thereby simultaneously realizing the benefits of parameter sparsity and decoding parallelism. By introducing delayed caching for decoded tokens and limited expert activation for cold tokens, TEAM reduces the number of activated experts by 35–39\% across all benchmarks, even after accounting for the additional experts introduced by multi-branch exploration. Moreover, speculative exploration for hot tokens further enhances decoding parallelism, increasing the number of accepted tokens per iteration by 1.49–1.74×. Benefiting from both higher sparsity and increased parallelism, TEAM requires on average only 6.92 activated experts to decode a single token, which is even lower than the nominal routing cost of 8 experts per token. As a result, TEAM achieves an average speedup of 1.94×, with a peak speedup of up to 2.2× on the HumanEval benchmark.

\subsection{Ablation Study and Hyperparameter Sensitivity}
\begin{figure}[t]
  \vskip 0.1in
  \begin{center}
    \centerline{\includegraphics[width=\columnwidth]{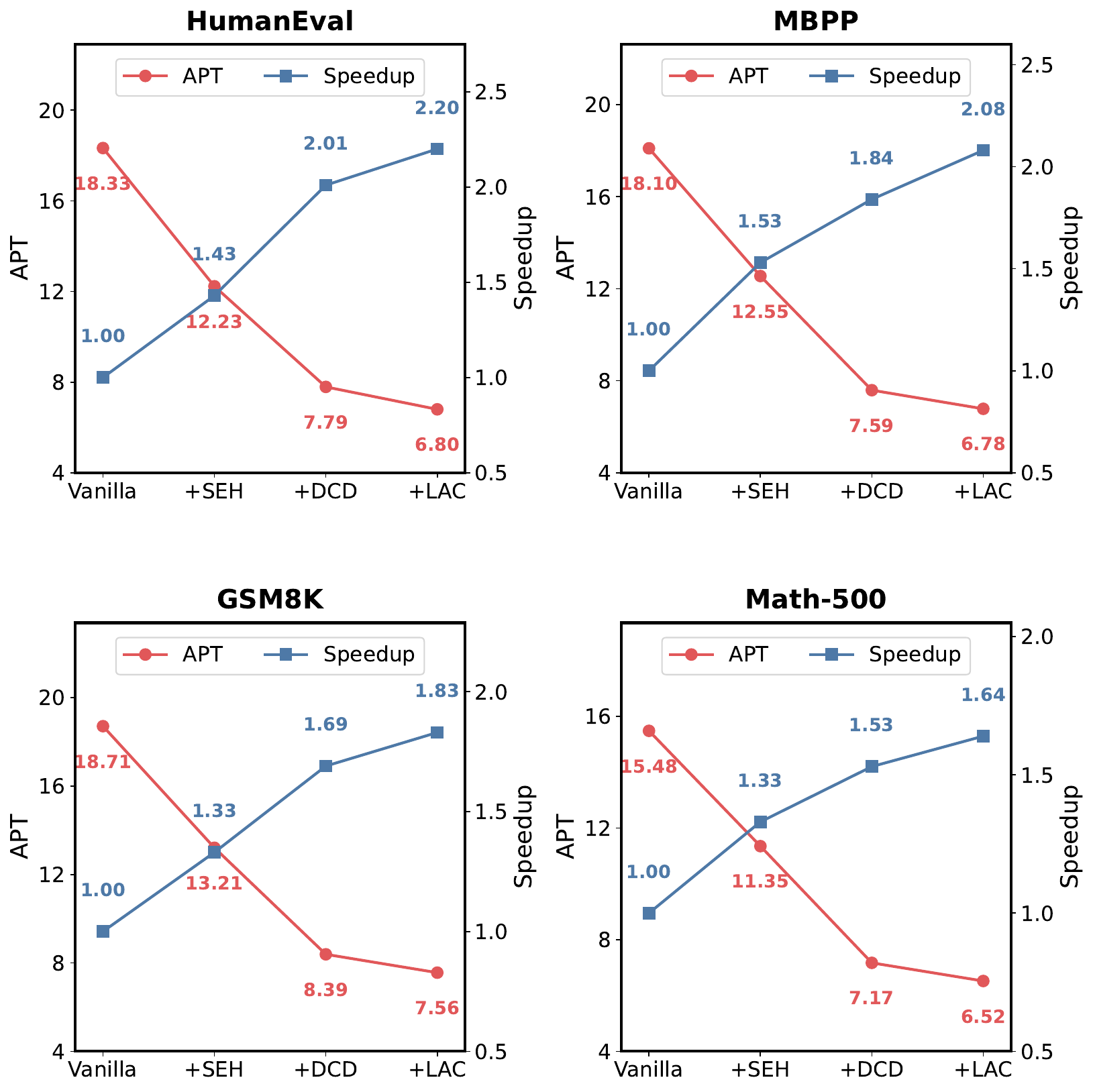}}
    \caption{
      Ablation study on the Activated experts Per decoded Token (APT) and speedup compared to the vanilla model.
    }
    \label{Fig5}
  \end{center}
\end{figure}

To assess the contribution of TEAM to accelerating inference in MoE dLLM, we progressively integrate its core techniques into the vanilla model. As illustrated in~\cref{Fig5}, we evaluate both the average number of activated experts required to decode a single token and the corresponding speedup throughout this process. 

The ablation study show that the introduction of Speculative Exploration for Hot Tokens (SEH) substantially reduces the number of activated experts per decoded token, validating its effectiveness. By decreasing the number of decoding steps, SEH increases the token acceptance rate per forward pass, while incuring only a marginal increase in additional expert activations thanks to the similarity across multiple branches. Delayed Caching for Decoded Tokens (DCD) further eliminates a large fraction of expert activations triggered by already decoded tokens that are irrelevant to the decoding process itself and only provide contextual guidance. Finally, Limited Activation for Cold Tokens (LAC) strictly confines expert activation to the subset responsible for newly decoded tokens and hot tokens. This design further reduces the number of activated experts per decoded token and yields additional speedup, resulting in the highest overall decoding efficiency among all benchmarks.

Furthermore, we will discuss the hyperparameter selection in our proposed TEAM.

A key factor in accelerating MoE dLLMs with TEAM is the criterion used to classify masked tokens into hot and cold categories, which is governed by a confidence threshold $\tau_h$ and a distance constraint $L_h$. We study the impact of different combinations of these two hyperparameters on model accuracy and the number of activated experts per forward pass on two code generation benchmarks, HumanEval and MBPP, as summarized in~\cref{hyperparam_sensitivity}. As $\tau_h$ increases up to 0.7 and $L_h$ decreases to 3, the model preserves its accuracy while the number of activated experts decreases monotonically, indicating improved efficiency. Further narrowing the set of tokens identified as hot leads to noticeable performance degradation. Consequently, the configuration $\tau_h=0.7$ and $L_h=3$ achieves the best accuracy-efficiency trade-off and is adopted as the default setting in TEAM.

\subsection{Analysis of Key Design Choices in TEAM}
\textbf{Refresh-free Caching for Decoded Tokens.} As discussed in Section~\ref{sec:TEAM methodology}, caching decoded tokens at every iteration like dKV-Cache yields limited acceleration for dLLMs that adopt the block diffusion paradigm. However, reducing expert activations through caching is critical to the efficiency of MoE dLLMs. Unlike prior work that periodically refreshes cached representations, we observe that for dLLMs initialized from autoregression, the representations of previously decoded tokens remain highly stable across diffusion iterations, rendering cache refresh unnecessary. As shown in~\cref{refresh_comparison}, we evaluate different refresh intervals, including recomputing all tokens every 4 steps, every 8 steps, or disabling refresh entirely within a block. We find no evidence that eliminating cache refresh leads to a noticeable degradation in model performance. In fact, the Refresh-free configuration even achieves slightly higher accuracy than Refresh-8. In contrast, the efficiency gains from removing refresh operations are substantial, making refresh-free caching a clear advantage for MoE dLLM inference.

\begin{figure}[t]
  \vskip 0.1in
  \begin{center}
    \centerline{\includegraphics[width=\columnwidth]{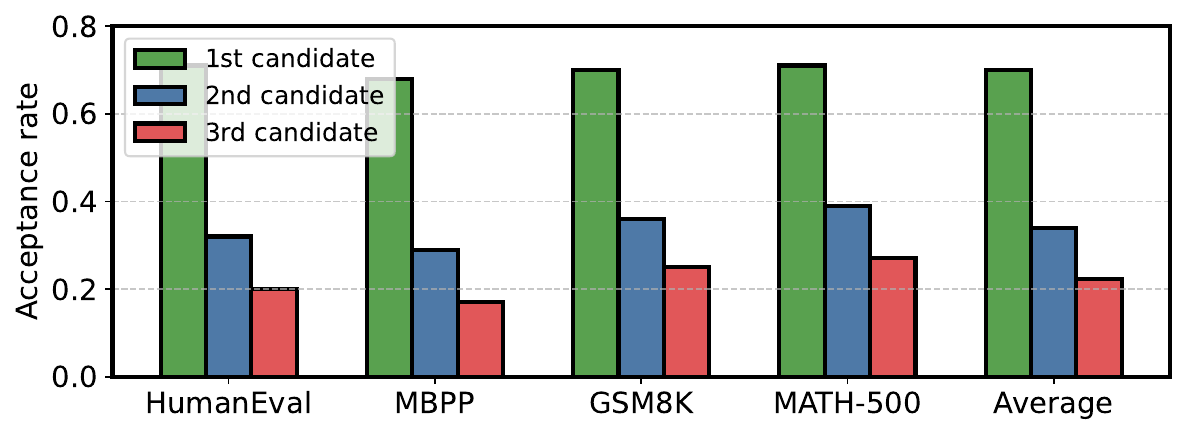}}
    \caption{
      Acceptance probability of the top3-confidence candidate tokens in the subsequent iteration in SEH.
    }
    \label{Fig6}
  \end{center}
\end{figure}

\textbf{Aligned Token Candidates in SEH.} To fully exploit the idle compute resources on GPU platforms, we introduce speculative exploration through additional acceptance of candidate tokens. Specifically, three extra branches are constructed by additionally accepting the 1st-confidence token, the 2nd-confidence token, and both tokens simultaneously, even when their confidence scores do not exceed the acceptance threshold. We argue that this design is preferable to further accepting the 3rd-confidence token. To justify this choice, we analyze the acceptance probability of the top3-confidence candidate tokens in the subsequent iteration, as illustrated in~\cref{Fig6}. The 2nd candidate has a relatively low probability of being directly accepted, while the 3rd one is rarely accepted. This indicates that constructing more diverse branches is inefficient. In contrast, aligned token combinations enable chained verification, thereby improving efficiency.

\textbf{The Shared Expert Subset for Masked Tokens.} A key observation of our proposed TEAM is that a specific subset of experts dominates the decoding process for most masked tokens, while the remaining experts contribute only marginally or are rarely activated. We further investigate the underlying reason for this phenomenon and provide evidence to support it. The dominance arises from the high similarity among model inputs for masked tokens. All masked tokens share the same [M] symbol and are mapped to identical token embeddings. The only source of variation comes from positional encodings, which are relatively minor within a block.As shown in~\cref{cosine_similarity}, we compute the average pairwise cosine similarity of hidden states among all masked tokens within a block at each decoding layer, and then report the highest and the average similarity in all layers. Such a high degree of similarity indicates that this homogeneity persists throughout the entire decoding process, leading to similar routing decisions.

\begin{table}[t]
    \centering
    % \vspace{0.1in}
    \caption{Highest and average pairwise cosine similarity of hidden states among all masked tokens within a block across different layers.}
    % \vspace{0.1in}
    \label{cosine_similarity}
    \small
    \setlength{\tabcolsep}{6pt}
    \begin{tabular}{ccccc}
        \toprule
         & \textbf{HumanEval} & \textbf{MBPP} & \textbf{GSM8K} & \textbf{MATH-500} \\
        \midrule
        Highest & 0.99 & 0.98 & 0.99 & 0.99 \\
        Average & 0.86 & 0.84 & 0.86 & 0.86 \\
        \bottomrule
    \end{tabular}
    \vspace{0.1in}
\end{table}

\section{Conclusion}
In this paper, we propose~\textbf{\method}, a framework for accelerating MoE diffusion language models through temporal and spatial consistency guided expert activation. Through a systematic analysis of temporal consistency across denoising iterations and spatial consistency across token positions,~\method~introduces three complementary expert activation and decoding strategies tailored to different subsets of tokens within each decoding block, enabling more tokens to be decoded with fewer activated experts. Compared to vanilla inference,~\method~achieves up to a 2.2× speedup, demonstrating an efficient and practical integration of dLLM and MoE architectures. 

\section*{Acknowledgments}
This work was supported in part by the National Key Research and Development Program under Grant 2024YFB4505004, in part by NSFC under Grant 62495102, Grant 92464104, and Grant 62341407, in part by Beijing Municipal Science and Technology Program under Grant Z241100004224015, in part by 111 Project under Grant B18001.

\section*{Impact Statement}
This paper presents work whose goal is to advance the field of Machine
Learning. There are many potential societal consequences of our work, none
which we feel must be specifically highlighted here.


\bibliography{references}

@article{nie2025large,
  title={Large language diffusion models},
  author={Nie, Shen and Zhu, Fengqi and You, Zebin and Zhang, Xiaolu and Ou, Jingyang and Hu, Jun and Zhou, Jun and Lin, Yankai and Wen, Ji-Rong and Li, Chongxuan},
  journal={arXiv preprint arXiv:2502.09992},
  year={2025}
}

@article{ye2025dream,
  title={Dream 7b: Diffusion large language models},
  author={Ye, Jiacheng and Xie, Zhihui and Zheng, Lin and Gao, Jiahui and Wu, Zirui and Jiang, Xin and Li, Zhenguo and Kong, Lingpeng},
  journal={arXiv preprint arXiv:2508.15487},
  year={2025}
}

@article{khanna2025mercury,
  title={Mercury: Ultra-fast language models based on diffusion},
  author={Khanna, Samar and Kharbanda, Siddhant and Li, Shufan and Varma, Harshit and Wang, Eric and Birnbaum, Sawyer and Luo, Ziyang and Miraoui, Yanis and Palrecha, Akash and Ermon, Stefano and others},
  journal={arXiv preprint arXiv:2506.17298},
  year={2025}
}

@article{wang2025diffusion,
  title={Diffusion llms can do faster-than-ar inference via discrete diffusion forcing},
  author={Wang, Xu and Xu, Chenkai and Jin, Yijie and Jin, Jiachun and Zhang, Hao and Deng, Zhijie},
  journal={arXiv preprint arXiv:2508.09192},
  year={2025}
}

@article{wu2025fast,
  title={Fast-dllm v2: Efficient block-diffusion llm},
  author={Wu, Chengyue and Zhang, Hao and Xue, Shuchen and Diao, Shizhe and Fu, Yonggan and Liu, Zhijian and Molchanov, Pavlo and Luo, Ping and Han, Song and Xie, Enze},
  journal={arXiv preprint arXiv:2509.26328},
  year={2025}
}

@article{fu2025efficient,
  title={Efficient-dlm: From autoregressive to diffusion language models, and beyond in speed},
  author={Fu, Yonggan and Whalen, Lexington and Ye, Zhifan and Dong, Xin and Diao, Shizhe and Liu, Jingyu and Wu, Chengyue and Zhang, Hao and Xie, Enze and Han, Song and others},
  journal={arXiv preprint arXiv:2512.14067},
  year={2025}
}

@article{liu2025wedlm,
  title={WeDLM: Reconciling Diffusion Language Models with Standard Causal Attention for Fast Inference},
  author={Liu, Aiwei and He, Minghua and Zeng, Shaoxun and Zhang, Sijun and Zhang, Linhao and Wu, Chuhan and Jia, Wei and Liu, Yuan and Zhou, Xiao and Zhou, Jie},
  journal={arXiv preprint arXiv:2512.22737},
  year={2025}
}

@article{shazeer2017outrageously,
  title={Outrageously large neural networks: The sparsely-gated mixture-of-experts layer},
  author={Shazeer, Noam and Mirhoseini, Azalia and Maziarz, Krzysztof and Davis, Andy and Le, Quoc and Hinton, Geoffrey and Dean, Jeff},
  journal={arXiv preprint arXiv:1701.06538},
  year={2017}
}

@article{jiang2024mixtral,
  title={Mixtral of experts},
  author={Jiang, Albert Q and Sablayrolles, Alexandre and Roux, Antoine and Mensch, Arthur and Savary, Blanche and Bamford, Chris and Chaplot, Devendra Singh and Casas, Diego de las and Hanna, Emma Bou and Bressand, Florian and others},
  journal={arXiv preprint arXiv:2401.04088},
  year={2024}
}

@article{qwen3,
    title={Qwen3 Technical Report}, 
    author={An Yang and Anfeng Li and Baosong Yang and Beichen Zhang and Binyuan Hui and Bo Zheng and Bowen Yu and Chang Gao and Chengen Huang and Chenxu Lv and Chujie Zheng and Dayiheng Liu and Fan Zhou and Fei Huang and Feng Hu and Hao Ge and Haoran Wei and Huan Lin and Jialong Tang and Jian Yang and Jianhong Tu and Jianwei Zhang and Jianxin Yang and Jiaxi Yang and Jing Zhou and Jingren Zhou and Junyang Lin and Kai Dang and Keqin Bao and Kexin Yang and Le Yu and Lianghao Deng and Mei Li and Mingfeng Xue and Mingze Li and Pei Zhang and Peng Wang and Qin Zhu and Rui Men and Ruize Gao and Shixuan Liu and Shuang Luo and Tianhao Li and Tianyi Tang and Wenbiao Yin and Xingzhang Ren and Xinyu Wang and Xinyu Zhang and Xuancheng Ren and Yang Fan and Yang Su and Yichang Zhang and Yinger Zhang and Yu Wan and Yuqiong Liu and Zekun Wang and Zeyu Cui and Zhenru Zhang and Zhipeng Zhou and Zihan Qiu},
    journal = {arXiv preprint arXiv:2505.09388},
    year={2025}
}

@article{liu2024deepseek,
  title={Deepseek-v3 technical report},
  author={Liu, Aixin and Feng, Bei and Xue, Bing and Wang, Bingxuan and Wu, Bochao and Lu, Chengda and Zhao, Chenggang and Deng, Chengqi and Zhang, Chenyu and Ruan, Chong and others},
  journal={arXiv preprint arXiv:2412.19437},
  year={2024}
}

@article{comanici2025gemini,
  title={Gemini 2.5: Pushing the frontier with advanced reasoning, multimodality, long context, and next generation agentic capabilities},
  author={Comanici, Gheorghe and Bieber, Eric and Schaekermann, Mike and Pasupat, Ice and Sachdeva, Noveen and Dhillon, Inderjit and Blistein, Marcel and Ram, Ori and Zhang, Dan and Rosen, Evan and others},
  journal={arXiv preprint arXiv:2507.06261},
  year={2025}
}

@article{cheng2025sdar,
  title={Sdar: A synergistic diffusion-autoregression paradigm for scalable sequence generation},
  author={Cheng, Shuang and Bian, Yihan and Liu, Dawei and Zhang, Linfeng and Yao, Qian and Tian, Zhongbo and Wang, Wenhai and Guo, Qipeng and Chen, Kai and Qi, Biqing and others},
  journal={arXiv preprint arXiv:2510.06303},
  year={2025}
}

@misc{bie2025llada20scalingdiffusionlanguage,
      title={LLaDA2.0: Scaling Up Diffusion Language Models to 100B}, 
      author={Tiwei Bie and Maosong Cao and Kun Chen and Lun Du and Mingliang Gong and Zhuochen Gong and Yanmei Gu and Jiaqi Hu and Zenan Huang and Zhenzhong Lan and Chengxi Li and Chongxuan Li and Jianguo Li and Zehuan Li and Huabin Liu and Ling Liu and Guoshan Lu and Xiaocheng Lu and Yuxin Ma and Jianfeng Tan and Lanning Wei and Ji-Rong Wen and Yipeng Xing and Xiaolu Zhang and Junbo Zhao and Da Zheng and Jun Zhou and Junlin Zhou and Zhanchao Zhou and Liwang Zhu and Yihong Zhuang},
      year={2025},
      eprint={2512.15745},
      archivePrefix={arXiv},
      primaryClass={cs.LG},
      url={https://arxiv.org/abs/2512.15745}, 
}

@article{dinfer,
  title={dinfer: An efficient inference framework for diffusion language models},
  author={Ma, Yuxin and Du, Lun and Wei, Lanning and Chen, Kun and Xu, Qian and Wang, Kangyu and Feng, Guofeng and Lu, Guoshan and Liu, Lin and Qi, Xiaojing and others},
  journal={arXiv preprint arXiv:2510.08666},
  year={2025}
}

@article{team2025every,
  title={Every Activation Boosted: Scaling General Reasoner to 1 Trillion Open Language Foundation},
  author={Team, Ling and Li, Ang and Liu, Ben and Hu, Binbin and Li, Bing and Zeng, Bingwei and Ye, Borui and Tang, Caizhi and Tian, Changxin and Huang, Chao and others},
  journal={arXiv preprint arXiv:2510.22115},
  year={2025}
}

@article{achiam2023gpt,
  title={Gpt-4 technical report},
  author={Achiam, Josh and Adler, Steven and Agarwal, Sandhini and Ahmad, Lama and Akkaya, Ilge and Aleman, Florencia Leoni and Almeida, Diogo and Altenschmidt, Janko and Altman, Sam and Anadkat, Shyamal and others},
  journal={arXiv preprint arXiv:2303.08774},
  year={2023}
}

@article{grattafiori2024llama,
  title={The llama 3 herd of models},
  author={Grattafiori, Aaron and Dubey, Abhimanyu and Jauhri, Abhinav and Pandey, Abhinav and Kadian, Abhishek and Al-Dahle, Ahmad and Letman, Aiesha and Mathur, Akhil and Schelten, Alan and Vaughan, Alex and others},
  journal={arXiv preprint arXiv:2407.21783},
  year={2024}
}

@article{li2025refusion,
  title={ReFusion: A Diffusion Large Language Model with Parallel Autoregressive Decoding},
  author={Li, Jia-Nan and Guan, Jian and Wu, Wei and Li, Chongxuan},
  journal={arXiv preprint arXiv:2512.13586},
  year={2025}
}

@article{tian2025next,
  title={From Next-Token to Next-Block: A Principled Adaptation Path for Diffusion LLMs},
  author={Tian, Yuchuan and Liang, Yuchen and Sun, Jiacheng and Zhang, Shuo and Yang, Guangwen and Shu, Yingte and Fang, Sibo and Guo, Tianyu and Han, Kai and Xu, Chao and others},
  journal={arXiv preprint arXiv:2512.06776},
  year={2025}
}

@inproceedings{arriolablock,
  title={Block Diffusion: Interpolating Between Autoregressive and Diffusion Language Models},
  author={Arriola, Marianne and Gokaslan, Aaron and Chiu, Justin T and Yang, Zhihan and Qi, Zhixuan and Han, Jiaqi and Sahoo, Subham Sekhar and Kuleshov, Volodymyr},
  booktitle={The Thirteenth International Conference on Learning Representations},
  year={2025}
}

@inproceedings{
    gong2025scaling,
    title={Scaling Diffusion Language Models via Adaptation from Autoregressive Models},
    author={Shansan Gong and Shivam Agarwal and Yizhe Zhang and Jiacheng Ye and Lin Zheng and Mukai Li and Chenxin An and Peilin Zhao and Wei Bi and Jiawei Han and Hao Peng and Lingpeng Kong},
    booktitle={The Thirteenth International Conference on Learning Representations},
    year={2025},
    url={https://openreview.net/forum?id=j1tSLYKwg8}
}

@article{zhu2025llada,
  title={Llada-moe: A sparse moe diffusion language model},
  author={Zhu, Fengqi and You, Zebin and Xing, Yipeng and Huang, Zenan and Liu, Lin and Zhuang, Yihong and Lu, Guoshan and Wang, Kangyu and Wang, Xudong and Wei, Lanning and others},
  journal={arXiv preprint arXiv:2509.24389},
  year={2025}
}

@article{liu2025dllm,
  title={dllm-cache: Accelerating diffusion large language models with adaptive caching},
  author={Liu, Zhiyuan and Yang, Yicun and Zhang, Yaojie and Chen, Junjie and Zou, Chang and Wei, Qingyuan and Wang, Shaobo and Zhang, Linfeng},
  journal={arXiv preprint arXiv:2506.06295},
  year={2025}
}

@article{ma2025dkv,
  title={dkv-cache: The cache for diffusion language models},
  author={Ma, Xinyin and Yu, Runpeng and Fang, Gongfan and Wang, Xinchao},
  journal={arXiv preprint arXiv:2505.15781},
  year={2025}
}

@article{chen2025dpad,
  title={Dpad: Efficient diffusion language models with suffix dropout},
  author={Chen, Xinhua and Huang, Sitao and Guo, Cong and Wei, Chiyue and He, Yintao and Zhang, Jianyi and Li, Hai and Chen, Yiran and others},
  journal={arXiv preprint arXiv:2508.14148},
  year={2025}
}

@article{song2025sparse,
  title={Sparse-dllm: Accelerating diffusion llms with dynamic cache eviction},
  author={Song, Yuerong and Liu, Xiaoran and Li, Ruixiao and Liu, Zhigeng and Huang, Zengfeng and Guo, Qipeng and He, Ziwei and Qiu, Xipeng},
  journal={arXiv preprint arXiv:2508.02558},
  year={2025}
}

@article{jiang2025d,
  title={d$^2$ Cache: Accelerating Diffusion-Based LLMs via Dual Adaptive Caching},
  author={Jiang, Yuchu and Cai, Yue and Luo, Xiangzhong and Fu, Jiale and Wang, Jiarui and Liu, Chonghan and Yang, Xu},
  journal={arXiv preprint arXiv:2509.23094},
  year={2025}
}

@article{qian2026d3llm,
  title={d3LLM: Ultra-Fast Diffusion LLM using Pseudo-Trajectory Distillation},
  author={Qian, Yu-Yang and Su, Junda and Hu, Lanxiang and Zhang, Peiyuan and Deng, Zhijie and Zhao, Peng and Zhang, Hao},
  journal={arXiv preprint arXiv:2601.07568},
  year={2026}
}

@article{gao2025self,
  title={Self Speculative Decoding for Diffusion Large Language Models},
  author={Gao, Yifeng and Ji, Ziang and Wang, Yuxuan and Qi, Biqing and Xu, Hanlin and Zhang, Linfeng},
  journal={arXiv preprint arXiv:2510.04147},
  year={2025}
}

@article{agrawal2025spiffy,
  title={Spiffy: Multiplying Diffusion LLM Acceleration via Lossless Speculative Decoding},
  author={Agrawal, Sudhanshu and Garrepalli, Risheek and Goel, Raghavv and Lee, Mingu and Lott, Christopher and Porikli, Fatih},
  journal={arXiv preprint arXiv:2509.18085},
  year={2025}
}

@article{wei2025orchestrating,
  title={Orchestrating Dual-Boundaries: An Arithmetic Intensity Inspired Acceleration Framework for Diffusion Language Models},
  author={Wei, Linye and Chen, Wenjue and Tang, Pingzhi and Guo, Xiaotian and Ye, Le and Wang, Runsheng and Li, Meng},
  journal={arXiv preprint arXiv:2511.21759},
  year={2025}
}

@article{wu2025free,
  title={Free Draft-and-Verification: Toward Lossless Parallel Decoding for Diffusion Large Language Models},
  author={Wu, Shutong and Zhang, Jiawei},
  journal={arXiv preprint arXiv:2510.00294},
  year={2025}
}

@article{chen2023accelerating,
  title={Accelerating large language model decoding with speculative sampling},
  author={Chen, Charlie and Borgeaud, Sebastian and Irving, Geoffrey and Lespiau, Jean-Baptiste and Sifre, Laurent and Jumper, John},
  journal={arXiv preprint arXiv:2302.01318},
  year={2023}
}

@inproceedings{leviathan2023fast,
  title={Fast inference from transformers via speculative decoding},
  author={Leviathan, Yaniv and Kalman, Matan and Matias, Yossi},
  booktitle={International Conference on Machine Learning},
  pages={19274--19286},
  year={2023},
  organization={PMLR}
}

@article{li2025sparse,
  title={Sparse-LaViDa: Sparse Multimodal Discrete Diffusion Language Models},
  author={Li, Shufan and Gu, Jiuxiang and Liu, Kangning and Lin, Zhe and Wei, Zijun and Grover, Aditya and Kuen, Jason},
  journal={arXiv preprint arXiv:2512.14008},
  year={2025}
}

@article{xu2025lopa,
  title={LoPA: Scaling dLLM Inference via Lookahead Parallel Decoding},
  author={Xu, Chenkai and Jin, Yijie and Li, Jiajun and Tu, Yi and Long, Guoping and Tu, Dandan and Hou, Tianqi and Yan, Junchi and Deng, Zhijie},
  journal={arXiv preprint arXiv:2512.16229},
  year={2025}
}

@article{humaneval,
  title={Evaluating Large Language Models Trained on Code},
  author={Mark Chen and Jerry Tworek and Heewoo Jun and Qiming Yuan and Henrique Ponde de Oliveira Pinto and Jared Kaplan and Harri Edwards and Yuri Burda and Nicholas Joseph and Greg Brockman and Alex Ray and Raul Puri and Gretchen Krueger and Michael Petrov and Heidy Khlaaf and Girish Sastry and Pamela Mishkin and Brooke Chan and Scott Gray and Nick Ryder and Mikhail Pavlov and Alethea Power and Lukasz Kaiser and Mohammad Bavarian and Clemens Winter and Philippe Tillet and Felipe Petroski Such and Dave Cummings and Matthias Plappert and Fotios Chantzis and Elizabeth Barnes and Ariel Herbert-Voss and William Hebgen Guss and Alex Nichol and Alex Paino and Nikolas Tezak and Jie Tang and Igor Babuschkin and Suchir Balaji and Shantanu Jain and William Saunders and Christopher Hesse and Andrew N. Carr and Jan Leike and Josh Achiam and Vedant Misra and Evan Morikawa and Alec Radford and Matthew Knight and Miles Brundage and Mira Murati and Katie Mayer and Peter Welinder and Bob McGrew and Dario Amodei and Sam McCandlish and Ilya Sutskever and Wojciech Zaremba},
  year={2021},
  eprint={2107.03374},
  archivePrefix={arXiv},
  primaryClass={cs.LG}
}

@article{austin2021program,
  title={Program synthesis with large language models},
  author={Austin, Jacob and Odena, Augustus and Nye, Maxwell and Bosma, Maarten and Michalewski, Henryk and Dohan, David and Jiang, Ellen and Cai, Carrie and Terry, Michael and Le, Quoc and others},
  journal={arXiv preprint arXiv:2108.07732},
  year={2021}
}

@article{cobbe2021training,
  title={Training verifiers to solve math word problems},
  author={Cobbe, Karl and Kosaraju, Vineet and Bavarian, Mohammad and Chen, Mark and Jun, Heewoo and Kaiser, Lukasz and Plappert, Matthias and Tworek, Jerry and Hilton, Jacob and Nakano, Reiichiro and others},
  journal={arXiv preprint arXiv:2110.14168},
  year={2021}
}

@inproceedings{lightman2023let,
  title={Let's verify step by step},
  author={Lightman, Hunter and Kosaraju, Vineet and Burda, Yuri and Edwards, Harrison and Baker, Bowen and Lee, Teddy and Leike, Jan and Schulman, John and Sutskever, Ilya and Cobbe, Karl},
  booktitle={The Twelfth International Conference on Learning Representations},
  year={2023}
}

@article{yang2026improving,
  title={Improving Sampling for Masked Diffusion Models via Information Gain},
  author={Yang, Kaisen and Teoh, Jayden and Yang, Kaicheng and Zhang, Yitong and Lamb, Alex},
  journal={arXiv preprint arXiv:2602.18176},
  year={2026}
}

@article{xu2026grouter,
  title={Grouter: Decoupling Routing from Representation for Accelerated MoE Training},
  author={Xu, Yuqi and Hu, Rizhen and Liu, Zihan and Sun, Mou and Yuan, Kun},
  journal={arXiv preprint arXiv:2603.06626},
  year={2026}
}

@article{feng2026dvoting,
  title={dvoting: Fast voting for dllms},
  author={Feng, Sicheng and Chen, Zigeng and Ma, Xinyin and Fang, Gongfan and Wang, Xinchao},
  journal={arXiv preprint arXiv:2602.12153},
  year={2026}
}

@article{zhu2025latent,
  title={Latent refinement decoding: Enhancing diffusion-based language models by refining belief states},
  author={Zhu, Qinglin and Yao, Yizhen and Zhao, Runcong and Xiang, Yanzheng and Saseendran, Amrutha and Jin, Chen and Teare, Philip and Liang, Bin and He, Yulan and Gui, Lin},
  journal={arXiv preprint arXiv:2510.11052},
  year={2025}
}

@article{yang2026efficientnav,
  title={Efficientnav: Towards on-device object-goal navigation with navigation map caching and retrieval},
  author={Yang, Zebin and Zheng, Sunjian and Xie, Tong and Xu, Tianshi and Yu, Bo and Wang, Fan and Tang, Jie and Liu, Shaoshan and Li, Meng},
  journal={Advances in Neural Information Processing Systems},
  volume={38},
  pages={4286--4312},
  year={2026}
}

@inproceedings{zhu2025pathology,
  title={Pathology-Aware Prototype Evolution via LLM-Driven Semantic Disambiguation for Multicenter Diabetic Retinopathy Diagnosis},
  author={Zhu, Chunzheng and Lin, Yangfang and Shao, Jialin and Lin, Jianxin and Wang, Yijun},
  booktitle={Proceedings of the 33rd ACM International Conference on Multimedia},
  pages={9196--9205},
  year={2025}
}

@article{ho2020denoising,
  title={Denoising diffusion probabilistic models},
  author={Ho, Jonathan and Jain, Ajay and Abbeel, Pieter},
  journal={Advances in neural information processing systems},
  volume={33},
  pages={6840--6851},
  year={2020}
}

@inproceedings{yan2025entropy,
  title={Entropy-Adaptive Diffusion Policy Optimization with Dynamic Step Alignment},
  author={Yan, RenYe and Cheng, Jikang and Gan, Yaozhong and Sun, Shikun and Wu, You and Yang, Yunfan and Ling, Liang and Lin, Jinlong and Zhu, Yeshuang and Zhou, Jie and others},
  booktitle={Proceedings of the IEEE/CVF International Conference on Computer Vision},
  pages={1924--1934},
  year={2025}
}

@article{yan2026less,
  title={Do Less, Achieve More: Do We Need Every-Step Optimization for RL Fine-tuning of Diffusion Models?},
  author={Yan, Renye and Cheng, Jikang and Sun, Shikun and Sun, Yi and Wu, You and Peng, Wei and Wang, Zongwei and Liang, Ling and Xing, Junliang and Cai, Yimao},
  journal={arXiv preprint arXiv:2605.15855},
  year={2026}
}

@article{zhu2025fmri2ges,
  title={fMRI2GES: Co-speech Gesture Reconstruction from fMRI Signal with Dual Brain Decoding Alignment},
  author={Zhu, Chunzheng and Shao, Jialin and Lin, Jianxin and Wang, Yijun and Wang, Jing and Tang, Jinhui and Li, Kenli},
  journal={IEEE Transactions on Circuits and Systems for Video Technology},
  year={2025},
  publisher={IEEE}
}

@inproceedings{huang2025hd,
  title={HD-MoE: Hybrid and Dynamic Parallelism for Mixture-of-Expert LLMs with 3D Near-Memory Processing},
  author={Huang, Haochen and Zhong, Shuzhang and Zhang, Zhe and Li, Shuangchen and Niu, Dimin and Zheng, Hongzhong and Wang, Runsheng and Li, Meng},
  booktitle={2025 IEEE/ACM International Conference On Computer Aided Design (ICCAD)},
  pages={1--9},
  year={2025},
  organization={IEEE}
}

@inproceedings{zhong2025hybrimoe,
  title={Hybrimoe: Hybrid cpu-gpu scheduling and cache management for efficient moe inference},
  author={Zhong, Shuzhang and Sun, Yanfan and Liang, Ling and Wang, Runsheng and Huang, Ru and Li, Meng},
  booktitle={2025 62nd ACM/IEEE Design Automation Conference (DAC)},
  pages={1--7},
  year={2025},
  organization={IEEE}
}

@article{yan2026pixel,
  title={Pixel-Space Diffusion Transformers},
  author={Yan, Renye and Cheng, Jikang and Wu, You and Liang, Ling and Peng, Wei and Vasilakos, Athanasios V and Zhao, Qingyu and Zhang, Yu and Adeli, Ehsan and Pohl, Kilian M and others},
  journal={arXiv preprint arXiv:2607.17585},
  year={2026}
}
\bibliographystyle{icml2026}

\end{document}